\documentclass[11pt,a4paper]{article}
\usepackage[hyperref]{eacl2021}
\usepackage{times}
\usepackage{latexsym}

\usepackage{placeins}
\usepackage{nicefrac}
\usepackage{latexsym}
\usepackage{multirow}
\usepackage{float}
\usepackage{booktabs}
\usepackage{graphicx}

\usepackage{pgfplots}
\pgfplotsset{compat=1.8}
\usetikzlibrary{patterns}
\usepackage{tikzsymbols}
\usepackage{graphicx}
\usepackage{fdsymbol}
\usepgfplotslibrary{patchplots}
\usepackage{filecontents}
\usepackage{enumitem}
\usepackage{booktabs}

\setlist[itemize]{leftmargin=*}
\setlist[enumerate]{leftmargin=*}
\usepackage{pifont}
\newcommand{\cmark}{\ding{51}}%
\newcommand{\xmark}{\ding{55}}%

\def\shortequals{\hspace{0.6mm}$=$\hspace{0.6mm}}

\pgfplotsset{compat=1.8,
    /pgfplots/xbar legend/.style={
    /pgfplots/legend image code/.code={%
       \draw[##1,/tikz/.cd,yshift=-0.25em]
        (0cm,0cm) rectangle (3pt,0.8em);},
   },
}
\usepackage{caption}
\usepackage{microtype}

\aclfinalcopy 

\title{Text Augmentation in a Multi-Task View}

\author{Jason Wei$^{\includegraphics[height=0.75em]{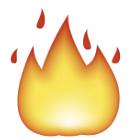}}$ \hspace{2.3mm} Chengyu Huang$^{\includegraphics[height=0.75em]{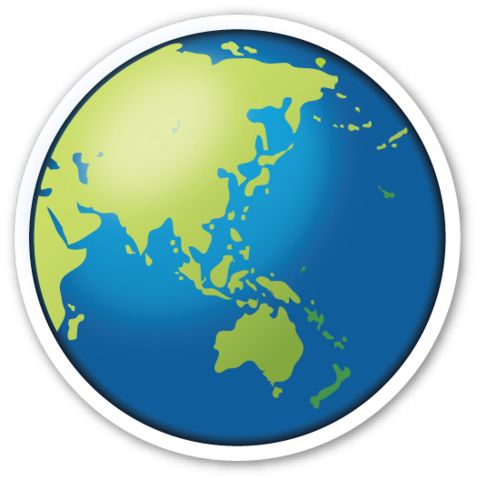}}$ \hspace{2.3mm} Shiqi Xu$^{\includegraphics[height=0.75em]{images/lit.png}}$ \hspace{2.3mm} Soroush Vosoughi$^{\Springtree}$ \\
  \includegraphics[height=0.75em]{images/lit.png}ProtagoLabs \\
    \includegraphics[height=0.75em]{images/world.png}International Monetary Fund \\
  {\Springtree}Dartmouth College \\
  \texttt{jason@protagolabs.com} 
  }

\date{}

\begin{document}
\maketitle
\begin{abstract}
Traditional data augmentation aims to increase the coverage of the input distribution by generating augmented examples that strongly resemble original samples in an online fashion where augmented examples dominate training.

In this paper, we propose an alternative perspective---a \textit{multi-task view} (MTV) of data augmentation---in which the primary task trains on original examples and the auxiliary task trains on augmented examples. 
In MTV data augmentation, both original and augmented samples are weighted substantively during training, relaxing the constraint that augmented examples must resemble original data and thereby allowing us to apply stronger levels of augmentation. 

In empirical experiments using four common data augmentation techniques on three benchmark text classification datasets, we find that the MTV leads to higher and more robust performance improvements than traditional augmentation.
\end{abstract}

\section{Introduction}

Most data augmentation techniques aim to generate augmented examples for training that are similar to original data.
In computer vision, operations such as flipping, cropping, and color jittering are both widely used and highly effective---it is self-evident that augmented examples closely resemble original data, and so we generate augmented data in an online fashion during each minibatch such that no original, unmodified examples are seen during training \cite{krizhevsky2012imagenet,BMVC2016_87,huang2017densely}.

In language, on the other hand, even slight modifications can cause significant semantic changes, and so it is not always clear whether augmented examples resemble original data.
Despite this uncertainty, many augmentation techniques in NLP still generate examples stochastically and ignore original data \cite{10.5555/2969239.2969312,sennrich-etal-2016-edinburgh,DBLP:conf/iclr/XieWLLNJN17,li-etal-2017-robust,kobayashi-2018-contextual,wang-etal-2018-switchout}.
When it is unclear whether augmented examples resemble original data---as is often the case---is it wise to neglect the original training data?

\begingroup
\begin{table}[t]
    \centering
    \small
    \begin{tabular}{p{7.3cm}}
    \underline{\textbf{Traditional Data Augmentation}} \\
    \begin{itemize}
        \vspace{-1.5em} \item \textbf{Intuition:} Increase coverage of input distribution by using augmented examples for training.
        \vspace{-0.8em} \item \textbf{Guideline:} Augmented examples should be similar to original data.
        \vspace{-0.8em} \item \textbf{Training:} Dominated by augmented examples that are generated stochastically.
    \end{itemize} \\
    \underline{\textbf{Multi-Task View (MTV) of Data Augmentation}} \\
    \begin{itemize}
        \vspace{-1.5em} \item \textbf{Intuition}: Auxiliary task of classifying augmented examples acts as regularization for the primary task of classifying original examples.
        \vspace{-0.8em} \item \textbf{Guideline}: It might be a good idea for augmented samples to resemble original data, but they can be anything that boosts performance.
        \vspace{-0.8em} \item \textbf{Training}: Both original and augmented data receive substantive weighting during training.
    \end{itemize}
    \end{tabular}
    \vspace{-1.3em} 
    \caption{
    Summary of traditional data augmentation versus MTV data augmentation. 
    }
    \label{tab:graphs}
\end{table}
\endgroup
Our paper questions this practice by proposing to include original data during training. Specifically, we make two contributions:
\begin{enumerate}
    \vspace{-0.5em} \item We propose a \textit{multi-task view of data augmentation} (MTV data augmentation), which trains on both original and augmented examples and therefore allows us to relax the constraint that augmented examples must resemble original data. The MTV facilitates augmentation using a higher strength parameter.
    \vspace{-0.5em} \item We show empirically that four common data augmentation techniques provide higher and more robust performance gains using the MTV compared with traditional augmentation. 
\end{enumerate}

\section{Traditional Data Augmentation\footnote{We closely follow the intuition and notation of Wang et al.~\shortcite{wang-etal-2018-switchout}}}
\label{sec:traditional}
\noindent \textbf{Situation.} During regular training, the canonical maximum likelihood objective minimizes the cost of the original training set $J_\textrm{O}$:
\vspace{-0.5em}
\[
J_\textrm{O}(\theta) = 
\mathbb{E}_{x,y\sim \pmb{\hat{\textrm{p}}}(X, Y)}
\left[ -\log \pmb{\textrm{p}}_\theta(y|x) \right] ,
\]
where $\pmb{\hat{\textrm{p}}}(X, Y)$ is the empirical distribution of training pairs $x, y$ and $\pmb{\textrm{p}}_\theta(y|x)$ is the parameterized model that we aim to learn (e.g., a neural network). 
As $\pmb{\hat{\textrm{p}}}(X, Y)$ is typically the observed data, it will likely have some mismatch with the true data distribution $\pmb{\textrm{p}}(X, Y)$. 
When the mismatch is dramatic---for instance, when $\pmb{\hat{\textrm{p}}}(X, Y)$ does not sufficiently cover the training space---model performance will likely suffer.

\vspace{0.2em} \noindent \textbf{Remedy.} In practice, we often use data augmentation to mitigate the inadequacy of $\pmb{\hat{\textrm{p}}}(X, Y)$ by providing additional training data. We generate an augmented distribution $\pmb{\textrm{q}}(\hat{X}, \hat{Y})$ and now minimize the cost of this augmented training set $J_\textrm{aug}$:
\vspace{-0.5em}
\[
J_\textrm{aug}(\theta) = 
\mathbb{E}_{x,y\sim \pmb{\textrm{q}}(\hat{X}, \hat{Y})}
\left[ -\log \pmb{\textrm{p}}_\theta(y|x) \right] .
\]
As we now optimize solely on $\pmb{\textrm{q}}(\hat{X}, \hat{Y})$, our goal is to find $(\hat{x}, \hat{y})$ pairs that are likely to fall in the true distribution $\pmb{\textrm{p}}$. 
Assuming the smoothness of $\pmb{\textrm{p}}$, similar $(x, y)$ pairs will have similar probabilities, and therefore if an augmented example is more similar to an observed example, it is more likely to be sampled under the true distribution. 
In other words, good augmented examples resemble the observed data, and we aim to find them. 
Conversely, if an augmented example diverges too far from any observed data, it is likely invalid and thus harmful for training; we don't want to train on these examples. 

The majority of prior work follows this framework of augmented examples resembling real data. 
As popular techniques, semantic noising substitutes tokens with synonyms
\cite{wang-yang-2015-thats,10.5555/2969239.2969312,li-etal-2017-robust};
Pervasive Dropout randomly removes words from the input sequence \cite{sennrich-etal-2016-edinburgh};
and SwitchOut (for machine translation) replaces some words in both source and target sentences with other words from their corresponding vocabularies \cite{wang-etal-2018-switchout}.

Moreover, most of these techniques perform augmentation on every training example in an online fashion, implicitly assuming that augmented examples so closely resemble original data that directly training on original examples is not even worth considering.
As we shall see in the next section, adding in these original examples during training might actually be a worthwhile idea.

\section{MTV Data Augmentation}
Multi-task optimization jointly trains on a primary task and one or more auxiliary tasks---the intuition is that requiring an algorithm to also learn an auxiliary task can act as better regularization than penalizing all complexity uniformly.
Prior work has found that multi-task models work particularly well when the tasks are similar, but can also improve performance even on unrelated tasks \cite{pmlr-v22-romera12,10.5555/3327546.3327586}.

We propose a multi-task view of data augmentation that has a primary task that optimizes regular training on original examples and an auxiliary task that optimizes training on augmented data. 
This MTV jointly optimizes the primary and auxiliary task(s) using a weighted cost function so that both original and augmented data receive substantial weight during training:
\vspace{-0.5em}
\[
J(\theta) = \gamma_{\textrm{O}}\cdot J_{\textrm{O}}(\theta) + \gamma_{\textrm{aug}}\cdot J_{\textrm{aug}}(\theta)\ ,
\]
where $\gamma_{\textrm{O}}$ is the weight of \textbf{o}riginal data and $\gamma_{\textrm{aug}}$ is the weight of \textbf{aug}mented data, and $\gamma_{\textrm{O}} + \gamma_{\textrm{aug}} = 1$.
In this context, observe that vanilla training uses $\gamma_{\textrm{O}}\shortequals1$ and $\gamma_{\textrm{aug}}\shortequals0$, and traditional data augmentation uses $\gamma_{\textrm{O}}\shortequals0$ and $\gamma_{\textrm{aug}}\shortequals1$.

The MTV gives us an important freedom that is not offered by the traditional data augmentation framework. 
Since traditional data augmentation only trains on augmented examples, performance suffers detrimentally when augmented data differs too much from the true distribution---therefore, most studies aim to generate augmented examples that resemble original data. 
MTV data augmentation, however, jointly trains on both original and augmented data, thereby allowing us to relax the constraint that original and augmented examples come from the same distribution. 
In fact, accepting that the original and augmented distributions might differ or could even be unrelated---as work in multi-task learning has done \cite{pmlr-v22-romera12,10.5555/3327546.3327586,pmlr-v9-rai10a}---liberates us to apply stronger levels of data augmentation, which, as we will demonstrate in the next section, leads to higher and more robust performance.

\section{Experiments}

This section compares multi-task view augmentation to traditional augmentation for various datasets and augmentation techniques. 

\subsection{Experimental Setup}

\noindent \textbf{Datasets.}~~We conduct experiments 
on three text classification tasks often used as benchmarks \cite{kim-2014-convolutional}: 
(1) Stanford Sentiment Treebank (\textbf{SST2}) \cite{socher-etal-2013-parsing} of movie reviews classified as positive/negative, 
(2) subjectivity/objectivity dataset (\textbf{SUBJ}) \cite{Pang:2004:SES:1218955.1218990}, where sentences are classified as either subjective or objective, and 
(3) question type dataset (\textbf{TREC}) \cite{Li:2002:LQC:1072228.1072378}, in which questions ask for either a description, entity, abbreviation, human, location, or number.

\vspace{0.35em} \noindent \textbf{Models and Experimental Procedures.}~~For text classification, we use BERT \cite{devlin-etal-2019-bert} (\texttt{bert-base-uncased} from HuggingFace) to extract features by averaging the last hidden states of the input tokens. 
To reduce the number of model hyperparameters and save computation time, we classify these features using a linear SVM trained for 1000 epochs.\footnote{This setup is not state-of-the-art but allows for experiments to be performed on CPU.}
Since training data size depends on the amount of augmented data, we adjust the number of training epochs so that all models receive the same number of updates.
All experiments are run for five random seeds.
Our baseline models without data augmentation achieved 84.5\%, 93.1\%, and 83.9\% accuracy respectively on the SST2, SUBJ, and TREC tasks. 

\vspace{0.35em} \noindent \textbf{Augmentation Techniques.}~~In this paper, we experiment with four simple and common data augmentation techniques studied in \citet{wei-zou-2019-eda}:
\textbf{(1) Token Substitution} \cite{10.5555/2969239.2969312} replaces words with WordNet \cite{miller1995wordnet} synonyms;
\textbf{(2) Pervasive Dropout} \cite{sennrich-etal-2016-edinburgh} applies word-level dropout;
\textbf{(3) Token Injection} \cite{wei-zou-2019-eda} insert a synonym of a random token in the sequence into a random position in that sequence;
\textbf{(4) Positional Shuffling} \cite{wei-zou-2019-eda} randomly chooses two tokens and swaps their positions.
For all four techniques, a parameter $\alpha$ indicates \textit{augmentation strength} by dictating how many perturbations are performed. 
For a given $\alpha$, we perform $n\shortequals\alpha$\hspace{0.7mm}$l$ perturbations, where $l$ is the sequence length. 

\begingroup
\setlength{\tabcolsep}{2.8pt}
\begin{table}[th]
    \centering
    \small
    \begin{tabular}{l | c c | c c c }
        Aug.~Technique & MTV & Best $\alpha$ & Avg.~Boost & ($\Delta_{\textrm{MTV}}$) \\
        \midrule
        \multirow{2}{*}{Token Substitution} & \xmark & 0.05 & 1.3\% & - \\
         & \cmark & 0.3 & 2.1\% & (+0.8\%) \\
        \midrule
        \multirow{2}{*}{Pervasive Dropout}  & \xmark & 0.1 & 1.8\% & - \\
         & \cmark & 0.4 & 2.5\% & (+0.7\%) \\
        \midrule
        \multirow{2}{*}{Token Injection} & \xmark & 0.05 & 0.7\% & - \\
         & \cmark & 0.5 & 2.2\% & (+1.5\%) \\
        \midrule
        \multirow{2}{*}{Positional Shuffling} & \xmark & 0.05 & 1.4\% & - \\
         & \cmark & 0.4 & 2.5\% & (+1.1\%) \\ 
    \end{tabular}
    \vspace{-1mm}
    \caption{
    Average performance boost on three text classification tasks for four augmentation techniques using the best-performing augmentation strength from $\alpha\in\{0.05, 0.1, 0.2, 0.3,0.4, 0.5\}$. 
    Traditional data augmentation works best at low $\alpha$, whereas MTV data augmentation provides the strongest performance for high $\alpha$. 
    $\Delta_{\textrm{MTV}}$ indicates additional boost from using the MTV compared with traditional augmentation.
    }
    \vspace{-3mm}
    \label{tab:summary-table}
\end{table}
\endgroup
\subsection{Stronger augmentation for more gains}
Table \ref{tab:summary-table} summarizes results for data augmentation in the MTV using $\gamma_\textrm{O}\shortequals\gamma_\textrm{aug}\shortequals0.5$ compared with traditional augmentation for the best-performing augmentation strength from $\alpha\in\{0.05, 0.1,0.2, 0.3,$ $0.4, 0.5\}$.
In the traditional framework, pervasive dropout had the strongest performance boost of 1.8\% using $\alpha\shortequals0.1$.
The MTV, however, allowed for stronger augmentation (i.e., $\alpha \geq 0.3$) that resulted in all four techniques to achieving boosts of more than 2.0\%. 

Perhaps strikingly, token injection and positional shuffling, which are less intuitive and not as commonly used as token substitution and pervasive dropout, achieve the strongest gains ($>1.0\%$) from using the MTV. 
One potential reason for this is that, compared with token substitution and pervasive dropout, token injection and positional shuffling are non-destructive in that they do not remove any of the original words, and so the nature of examples augmented at high $\alpha$ could be more conducive for the MTV.

\pgfplotsset{width=3.86cm,height=3.03cm,compat=1.9}
\pgfplotsset{every tick label/.append style={font=\footnotesize}}
\begin{figure*}[th]
\vspace{-1.3mm}
%
%
%
%
A:
\vspace{-2mm}
\begin{tikzpicture}
\begin{axis}[
    title={\textbf{SUBJ}},
    title style={yshift=-1.5ex,},
    ylabel={Boost (\%)},
    xmin=-0.03, xmax=0.53,
    ymin=-4.4, ymax=4.4,
    xtick={0, 0.2, 0.4},
    ytick={-4, -2, 0, 2, 4},
    xticklabels={,,},
    legend pos=outer north east,
    ymajorgrids=true,
    xmajorgrids=true,
    grid style=dashed,
    x label style={at={(axis description cs:0.5,-0.25)},anchor=north,font=\small},
    y label style={at={(axis description cs:-0.22,0.5)},anchor=south,font=\small},
]
\addplot[
    color=blue,
    mark=o,
    mark size=2.5pt,
    ]
    coordinates {
    (0.0,   0.0)
    (0.05,  1.0)
    (0.1,   0.8)
    (0.2,   1.3)
    (0.3,   1.0)
    (0.4,   0.9)
    (0.5,   1.0)
    };
\addplot[
    color=red,
    mark=triangle,
    mark size=3.5pt,
    ]
    coordinates {
    (0.0,   0.0)
    (0.05,  1.1)
    (0.1,   1.2)
    (0.2,   0.2)
    (0.3,   -0.6)
    (0.4,   -1.7)
    (0.5,   -2.8)
    };
\end{axis}
\end{tikzpicture}
\begin{tikzpicture}
\begin{axis}[
    title={\textbf{SST2}},
    title style={yshift=-1.5ex,},
    xmin=-0.03, xmax=0.53,
    ymin=-4.4, ymax=4.4,
    xtick={0, 0.2, 0.4},
    ytick={-4, -2, 0, 2, 4},
    xticklabels={,,},
    legend pos=outer north east,
    ymajorgrids=true,
    xmajorgrids=true,
    grid style=dashed,
    x label style={at={(axis description cs:0.5,-0.18)},anchor=north,font=\small},
    y label style={at={(axis description cs:-0.27,0.5)},anchor=south,font=\small},
]
\addplot[
    color=blue,
    mark=o,
    mark size=2.5pt,
    ]
    coordinates {
    (0.0,   0.0)
    (0.05,  0.4)
    (0.1,   0.3)
    (0.2,   0.3)
    (0.3,   0)
    (0.4,   -0.1)
    (0.5,   0.1)
    };
\addplot[
    color=red,
    mark=triangle,
    mark size=3.5pt,
    ]
    coordinates {
    (0.0,   0.0)
    (0.05,  0.1)
    (0.1,   -0.1)
    (0.2,   0)
    (0.3,   -0.8)
    (0.4,   -1.2)
    (0.5,   -1.3)
    };
\end{axis}
\end{tikzpicture}
\begin{tikzpicture}
\begin{axis}[
    title={\textbf{TREC}},
    title style={yshift=-1.5ex,},
    xmin=-0.03, xmax=0.53,
    ymin=-6.5, ymax=6.5,
    xtick={0, 0.2, 0.4},
    ytick={-6, -3, 0, 3, 6},
    xticklabels={,,},
    legend pos=outer north east,
    ymajorgrids=true,
    xmajorgrids=true,
    grid style=dashed,
    x label style={at={(axis description cs:0.5,-0.18)},anchor=north,font=\small},
    y label style={at={(axis description cs:-0.27,0.5)},anchor=south,font=\small},
]
\addplot[
    color=blue,
    mark=o,
    mark size=2.5pt,
    ]
    coordinates {
    (0.0,   0.0)
    (0.05,  5)
    (0.1,   3.4)
    (0.2,   4.4)
    (0.3,   5.3)
    (0.4,   3.7)
    (0.5,   4.6)
    };
\addplot[
    color=red,
    mark=triangle,
    mark size=3.5pt,
    ]
    coordinates {
    (0.0,   0.0)
    (0.05,  2.9)
    (0.1,   2.5)
    (0.2,   1.2)
    (0.3,   2.5)
    (0.4,   1.9)
    (0.5,   -3.7)
    };
\end{axis}
\end{tikzpicture}
\begin{tikzpicture}
\begin{axis}[
    title={\textbf{\textsc{Average}}},
    title style={yshift=-1.5ex,},
    xmin=-0.03, xmax=0.53,
    ymin=-4.4, ymax=4.4,
    xtick={0, 0.2, 0.4},
    ytick={-4, -2, 0, 2, 4},
    xticklabels={,,},
    legend pos=outer north east,
    ymajorgrids=true,
    xmajorgrids=true,
    grid style=dashed,
    x label style={at={(axis description cs:-0.5,-0.25)},anchor=north},
    y label style={at={(axis description cs:-0.22,0.5)},anchor=south,font=\small},
]
\addplot[
    color=blue,
    mark=o,
    mark size=2.5pt,
    ]
    coordinates {
    (0.0,   0.0)
    (0.05,  2.1)
    (0.1,   1.47)
    (0.2,   1.97)
    (0.3,   2.07)
    (0.4,   1.47)
    (0.5,   1.87)
    };
    \addlegendentry{MTV}
\addplot[
    color=red,
    mark=triangle,
    mark size=3.5pt,
    ]
    coordinates {
    (0.0,   0.0)
    (0.05,  1.33)
    (0.1,   1.17)
    (0.2,   0.43)
    (0.3,   0.33)
    (0.4,   -0.37)
    (0.5,   -2.63)
    };
    \addlegendentry{Traditional}
\end{axis}
\end{tikzpicture}
\newline
B:
\vspace{-2mm}
\begin{tikzpicture}
\begin{axis}[
    ylabel={Boost (\%)},
    xmin=-0.03, xmax=0.53,
    ymin=-4.4, ymax=4.4,
    xtick={0, 0.2, 0.4},
    ytick={-4, -2, 0, 2, 4},
    xticklabels={,,},
    legend pos=outer north east,
    ymajorgrids=true,
    xmajorgrids=true,
    grid style=dashed,
    x label style={at={(axis description cs:0.5,-0.25)},anchor=north,font=\small},
    y label style={at={(axis description cs:-0.22,0.5)},anchor=south,font=\small},
]
\addplot[
    color=blue,
    mark=o,
    mark size=2.5pt,
    ]
    coordinates {
    (0.0,   0.0)
    (0.05,  1.0)
    (0.1,   1.3)
    (0.2,   2.0)
    (0.3,   1.6)
    (0.4,   2.0)
    (0.5,   1.4)
    };
\addplot[
    color=red,
    mark=triangle,
    mark size=3.5pt,
    ]
    coordinates {
    (0.0,   0.0)
    (0.05,  0.6)
    (0.1,   1.9)
    (0.2,   1.5)
    (0.3,   1.4)
    (0.4,   1.2)
    (0.5,   0.0)
    };
\end{axis}
\end{tikzpicture}
\begin{tikzpicture}
\begin{axis}[
    xmin=-0.03, xmax=0.53,
    ymin=-4.4, ymax=4.4,
    xtick={0, 0.2, 0.4},
    ytick={-4, -2, 0, 2, 4},
    xticklabels={,,},
    legend pos=outer north east,
    ymajorgrids=true,
    xmajorgrids=true,
    grid style=dashed,
    x label style={at={(axis description cs:0.5,-0.18)},anchor=north,font=\small},
    y label style={at={(axis description cs:-0.27,0.5)},anchor=south,font=\small},
]
\addplot[
    color=blue,
    mark=o,
    mark size=2.5pt,
    ]
    coordinates {
    (0.0,   0.0)
    (0.05,  -0.1)
    (0.1,   0.3)
    (0.2,   0.2)
    (0.3,   0.2)
    (0.4,   -0.1)
    (0.5,   0.9)
    };
\addplot[
    color=red,
    mark=triangle,
    mark size=3.5pt,
    ]
    coordinates {
    (0.0,   0.0)
    (0.05,  0.2)
    (0.1,   0.9)
    (0.2,   -0.4)
    (0.3,   1.1)
    (0.4,   -0.6)
    (0.5,   -2.7)
    };
\end{axis}
\end{tikzpicture}
\begin{tikzpicture}
\begin{axis}[
    xmin=-0.03, xmax=0.53,
    ymin=-6.5, ymax=6.5,
    xtick={0, 0.2, 0.4},
    ytick={-6, -3, 0, 3, 6},
    xticklabels={,,},
    legend pos=outer north east,
    ymajorgrids=true,
    xmajorgrids=true,
    grid style=dashed,
    x label style={at={(axis description cs:0.5,-0.18)},anchor=north,font=\small},
    y label style={at={(axis description cs:-0.27,0.5)},anchor=south,font=\small},
]
\addplot[
    color=blue,
    mark=o,
    mark size=2.5pt,
    ]
    coordinates {
    (0.0,   0.0)
    (0.05,  2)
    (0.1,   3.1)
    (0.2,   3.8)
    (0.3,   5.1)
    (0.4,   5.8)
    (0.5,   5.1)
    };
\addplot[
    color=red,
    mark=triangle,
    mark size=3.5pt,
    ]
    coordinates {
    (0.0,   0.0)
    (0.05,  1.7)
    (0.1,   2.7)
    (0.2,   3.6)
    (0.3,   2.9)
    (0.4,   2)
    (0.5,   -4.8)
    };
\end{axis}
\end{tikzpicture}
\begin{tikzpicture}
\begin{axis}[
    title style={yshift=-1.5ex,},
    xmin=-0.03, xmax=0.53,
    ymin=-4.4, ymax=4.4,
    xtick={0, 0.2, 0.4},
    ytick={-4, -2, 0, 2, 4},
    xticklabels={,,},
    legend pos=outer north east,
    ymajorgrids=true,
    xmajorgrids=true,
    grid style=dashed,
    x label style={at={(axis description cs:0.5,-0.33)},anchor=north,font=\small},
    y label style={at={(axis description cs:-0.22,0.5)},anchor=south,font=\small},
]
\addplot[
    color=blue,
    mark=o,
    mark size=2.5pt,
    ]
    coordinates {
    (0.0,   0.0)
    (0.05,  0.93)
    (0.1,   1.53)
    (0.2,   1.97)
    (0.3,   2.27)
    (0.4,   2.53)
    (0.5,   2.43)
    };
\addplot[
    color=red,
    mark=triangle,
    mark size=3.5pt,
    ]
    coordinates {
    (0.0,   0.0)
    (0.05,  0.8)
    (0.1,   1.8)
    (0.2,   1.53)
    (0.3,   1.77)
    (0.4,   0.83)
    (0.5,   -2.53)
    };
\end{axis}
\end{tikzpicture}
\newline
C:
\vspace{-2mm}
\begin{tikzpicture}
\begin{axis}[
    ylabel={Boost (\%)},
    xmin=-0.03, xmax=0.53,
    ymin=-4.4, ymax=4.4,
    xtick={0, 0.2, 0.4},
    ytick={-4, -2, 0, 2, 4},
    xticklabels={,,},
    legend pos=outer north east,
    ymajorgrids=true,
    xmajorgrids=true,
    grid style=dashed,
    x label style={at={(axis description cs:0.5,-0.25)},anchor=north,font=\small},
    y label style={at={(axis description cs:-0.22,0.5)},anchor=south,font=\small},
]
\addplot[
    color=blue,
    mark=o,
    mark size=2.5pt,
    ]
    coordinates {
    (0.0,   0.0)
    (0.05,  1.0)
    (0.1,   0.6)
    (0.2,   1.5)
    (0.3,   1.6)
    (0.4,   1.1)
    (0.5,   0.9)
    };
\addplot[
    color=red,
    mark=triangle,
    mark size=3.5pt,
    ]
    coordinates {
    (0.0,   0.0)
    (0.05,  1.0)
    (0.1,   0.3)
    (0.2,   -0.3)
    (0.3,   0.1)
    (0.4,   -0.6)
    (0.5,   -0.8)
    };
\end{axis}
\end{tikzpicture}
\begin{tikzpicture}
\begin{axis}[
    xmin=-0.03, xmax=0.53,
    ymin=-4.4, ymax=4.4,
    xtick={0, 0.2, 0.4},
    ytick={-4, -2, 0, 2, 4},
    xticklabels={,,},
    legend pos=outer north east,
    ymajorgrids=true,
    xmajorgrids=true,
    grid style=dashed,
    x label style={at={(axis description cs:0.5,-0.18)},anchor=north,font=\small},
    y label style={at={(axis description cs:-0.27,0.5)},anchor=south,font=\small},
]
\addplot[
    color=blue,
    mark=o,
    mark size=2.5pt,
    ]
    coordinates {
    (0.0,   0.0)
    (0.05,  -0.2)
    (0.1,   0.0)
    (0.2,   -0.1)
    (0.3,   -0.2)
    (0.4,   0.3)
    (0.5,   0.6)
    };
\addplot[
    color=red,
    mark=triangle,
    mark size=3.5pt,
    ]
    coordinates {
    (0.0,   0.0)
    (0.05,  0)
    (0.1,   -0.2)
    (0.2,   -1.2)
    (0.3,   -1)
    (0.4,   -1.6)
    (0.5,   -1.8)
    };
\end{axis}
\end{tikzpicture}
\begin{tikzpicture}
\begin{axis}[
    xmin=-0.03, xmax=0.53,
    ymin=-6.5, ymax=6.5,
    xtick={0, 0.2, 0.4},
    ytick={-6, -3, 0, 3, 6},
    xticklabels={,,},
    legend pos=outer north east,
    ymajorgrids=true,
    xmajorgrids=true,
    grid style=dashed,
    x label style={at={(axis description cs:0.5,-0.18)},anchor=north,font=\small},
    y label style={at={(axis description cs:-0.27,0.5)},anchor=south,font=\small},
]
\addplot[
    color=blue,
    mark=o,
    mark size=2.5pt,
    ]
    coordinates {
    (0.0,   0.0)
    (0.05,  3.5)
    (0.1,   3.3)
    (0.2,   3.9)
    (0.3,   4.4)
    (0.4,   5.1)
    (0.5,   5.1)
    };
\addplot[
    color=red,
    mark=triangle,
    mark size=3.5pt,
    ]
    coordinates {
    (0.0,   0.0)
    (0.05,  1.1)
    (0.1,   0.9)
    (0.2,   2.4)
    (0.3,   2.7)
    (0.4,   -0.7)
    (0.5,   1.9)
    };
\end{axis}
\end{tikzpicture}
\begin{tikzpicture}
\begin{axis}[
    xmin=-0.03, xmax=0.53,
    ymin=-4.4, ymax=4.4,
    xtick={0, 0.2, 0.4},
    ytick={-4, -2, 0, 2, 4},
    xticklabels={,,},
    legend pos=outer north east,
    ymajorgrids=true,
    xmajorgrids=true,
    grid style=dashed,
    x label style={at={(axis description cs:0.5,-0.18)},anchor=north,font=\small},
    y label style={at={(axis description cs:-0.22,0.5)},anchor=south,font=\small},
]
\addplot[
    color=blue,
    mark=o,
    mark size=2.5pt,
    ]
    coordinates {
    (0.0,   0.0)
    (0.05,  1.4)
    (0.1,   1.27)
    (0.2,   1.73)
    (0.3,   1.9)
    (0.4,   2.13)
    (0.5,   2.17)
    };
\addplot[
    color=red,
    mark=triangle,
    mark size=3.5pt,
    ]
    coordinates {
    (0.0,   0.0)
    (0.05,  0.67)
    (0.1,   0.3)
    (0.2,   0.27)
    (0.3,   0.57)
    (0.4,   -1.0)
    (0.5,   -1.53)
    };
\end{axis}
\end{tikzpicture}
\newline
D:
\vspace{-2mm}
\begin{tikzpicture}
\begin{axis}[
    xlabel={Strength $\alpha$},
    ylabel={Boost (\%)},
    xmin=-0.03, xmax=0.53,
    ymin=-4.4, ymax=4.4,
    xtick={0, 0.2, 0.4},
    ytick={-4, -2, 0, 2, 4},
    legend pos=outer north east,
    ymajorgrids=true,
    xmajorgrids=true,
    grid style=dashed,
    x label style={at={(axis description cs:0.5,-0.18)},anchor=north,font=\small},
    y label style={at={(axis description cs:-0.22,0.5)},anchor=south,font=\small},
]
\addplot[
    color=blue,
    mark=o,
    mark size=2.5pt,
    ]
    coordinates {
    (0.0,   0.0)
    (0.05,  1.2)
    (0.1,   1.8)
    (0.2,   1.3)
    (0.3,   1.8)
    (0.4,   1.5)
    (0.5,   1.2)
    };
\addplot[
    color=red,
    mark=triangle,
    mark size=3.5pt,
    ]
    coordinates {
    (0.0,   0.0)
    (0.05,  1.1)
    (0.1,   1.2)
    (0.2,   -1.2)
    (0.3,   -1.5)
    (0.4,   -2.3)
    (0.5,   -3.2)
    };
\end{axis}
\end{tikzpicture}
\begin{tikzpicture}
\begin{axis}[
    xlabel={Strength $\alpha$},
    xmin=-0.03, xmax=0.53,
    ymin=-4.4, ymax=4.4,
    xtick={0, 0.2, 0.4},
    ytick={-4, -2, 0, 2, 4},
    legend pos=outer north east,
    ymajorgrids=true,
    xmajorgrids=true,
    grid style=dashed,
    x label style={at={(axis description cs:0.5,-0.18)},anchor=north,font=\small},
    y label style={at={(axis description cs:-0.27,0.5)},anchor=south,font=\small},
]
\addplot[
    color=blue,
    mark=o,
    mark size=2.5pt,
    ]
    coordinates {
    (0.0,   0.0)
    (0.05,  0.0)
    (0.1,   -0.6)
    (0.2,   -0.1)
    (0.3,   0.1)
    (0.4,   0.6)
    (0.5,   0.1)
    };
\addplot[
    color=red,
    mark=triangle,
    mark size=3.5pt,
    ]
    coordinates {
    (0.0,   0.0)
    (0.05,  0.2)
    (0.1,   -0.4)
    (0.2,   -0.4)
    (0.3,   -0.5)
    (0.4,   -0.5)
    (0.5,   -2.6)
    };
\end{axis}
\end{tikzpicture}
\begin{tikzpicture}
\begin{axis}[
    xlabel={Strength $\alpha$},
    xmin=-0.03, xmax=0.53,
    ymin=-6.5, ymax=6.5,
    xtick={0, 0.2, 0.4},
    ytick={-6, -3, 0, 3, 6},
    legend pos=outer north east,
    ymajorgrids=true,
    xmajorgrids=true,
    grid style=dashed,
    x label style={at={(axis description cs:0.5,-0.18)},anchor=north,font=\small},
    y label style={at={(axis description cs:-0.27,0.5)},anchor=south,font=\small},
]
\addplot[
    color=blue,
    mark=o,
    mark size=2.5pt,
    ]
    coordinates {
    (0.0,   0.0)
    (0.05,  4.6)
    (0.1,   5.1)
    (0.2,   6)
    (0.3,   4.1)
    (0.4,   5.4)
    (0.5,   3.9)
    };
\addplot[
    color=red,
    mark=triangle,
    mark size=3.5pt,
    ]
    coordinates {
    (0.0,   0.0)
    (0.05,  3.1)
    (0.1,   1.8)
    (0.2,   3.8)
    (0.3,   1.3)
    (0.4,   0)
    (0.5,   1.5)
    };
\end{axis}
\end{tikzpicture}
\begin{tikzpicture}
\begin{axis}[
    xlabel={Strength $\alpha$},
    xmin=-0.03, xmax=0.53,
    ymin=-4.4, ymax=4.4,
    xtick={0, 0.2, 0.4},
    ytick={-4, -2, 0, 2, 4},
    legend pos=outer north east,
    ymajorgrids=true,
    xmajorgrids=true,
    grid style=dashed,
    x label style={at={(axis description cs:0.5,-0.18)},anchor=north,font=\small},
    y label style={at={(axis description cs:-0.22,0.5)},anchor=south,font=\small},
]
\addplot[
    color=blue,
    mark=o,
    mark size=2.5pt,
    ]
    coordinates {
    (0.0,   0.0)
    (0.05,  1.9)
    (0.1,   2.1)
    (0.2,   2.4)
    (0.3,   1.97)
    (0.4,   2.47)
    (0.5,   1.7)
    };
\addplot[
    color=red,
    mark=triangle,
    mark size=3.5pt,
    ]
    coordinates {
    (0.0,   0.0)
    (0.05,  1.43)
    (0.1,   0.83)
    (0.2,   0.7)
    (0.3,   -0.27)
    (0.4,   -0.1)
    (0.5,   -1.5)
    };
\end{axis}
\end{tikzpicture}
\vspace{3.0mm}
\caption{
Performance boosts on three tasks using the traditional and multi-task view (MTV) frameworks for four data augmentation techniques: Token Substitution \cite{10.5555/2969239.2969312} (A), Pervasive Dropout \cite{sennrich-etal-2016-edinburgh} (B), Token Injection \cite{wei-zou-2019-eda} (C), Positional Shuffling \cite{wei-zou-2019-eda} (D). 
In the traditional framework, improvements are largest when augmentation strength $\alpha$ is small, with performance deteriorating for large $\alpha$. 
The MTV, on the other hand, jointly optimizes for both original and augmented data, leveraging higher $\alpha$ to provide higher and more robust performance gains.
}
\vspace{-1.0mm}
\label{fig:graphs}
\end{figure*}
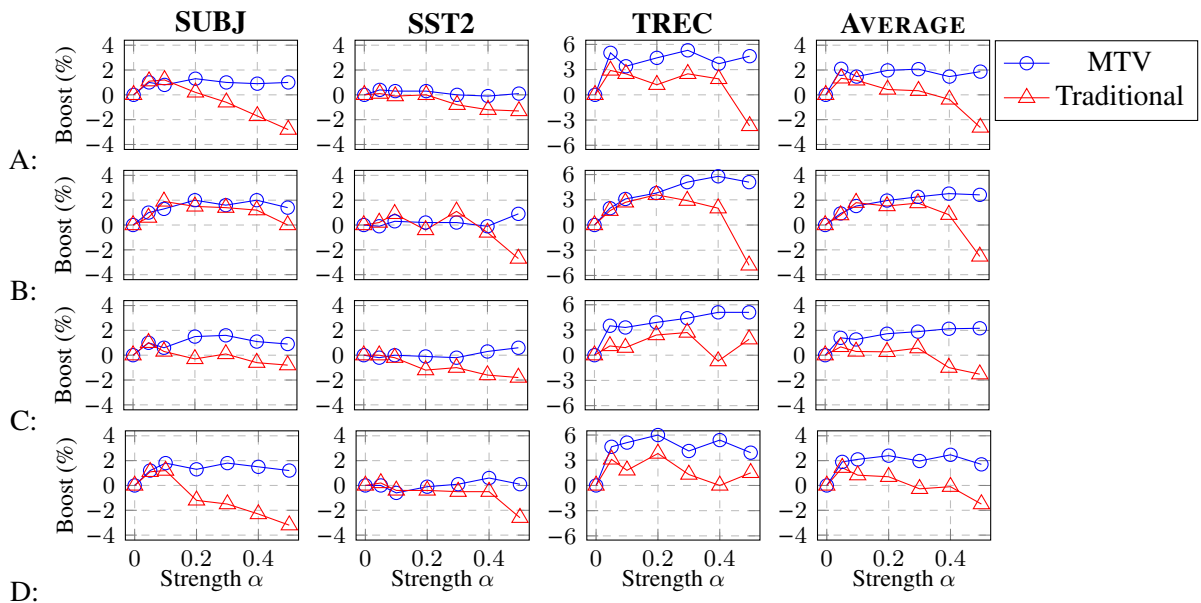
\subsection{More-robust gains at high $\alpha$}
When using data augmentation with high $\alpha$, high levels of noising are employed and augmented data are therefore more likely to diverge from their original examples. 
Figure \ref{fig:graphs} takes a closer look at how performance is affected by varying $\alpha$. 
Whereas traditional augmentation often negatively affected performance at high $\alpha$, the multi-task view, which jointly optimizes the original distribution, had robust performance gains at high augmentation strengths.

\begin{figure}[ht]
    \centering
    \includegraphics[width=\linewidth]{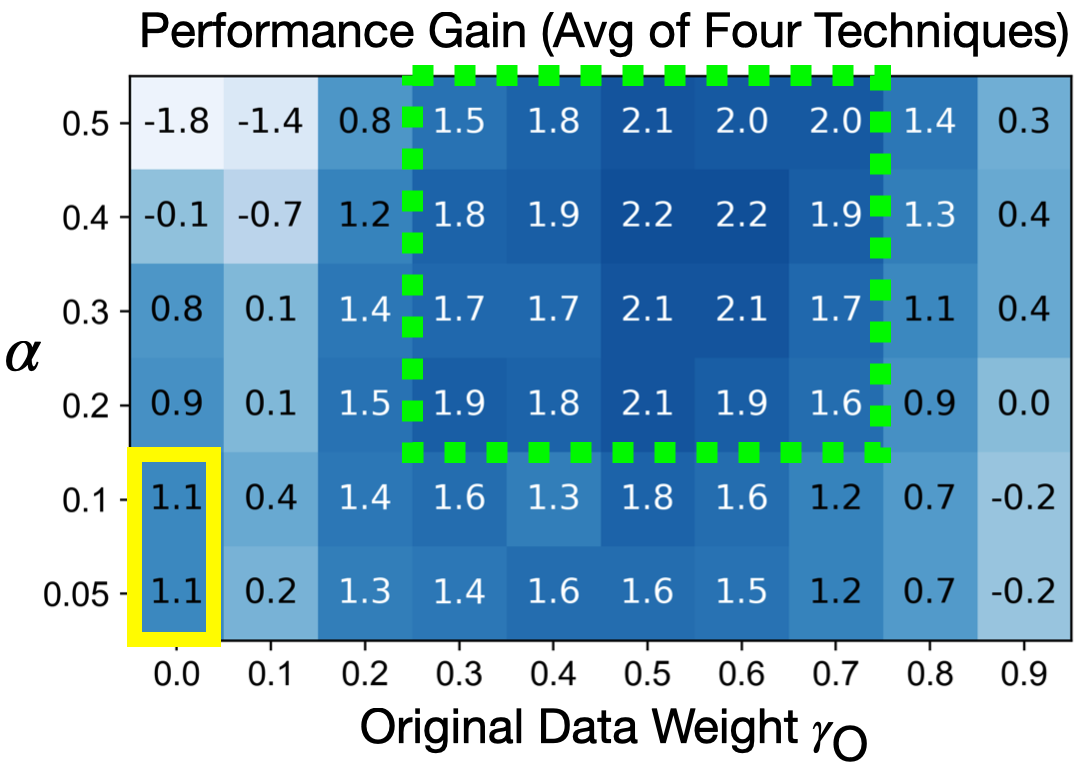}
    \vspace{-6.3mm}
    \caption{
    Performance boost (\%) with varying augmentation strengths $\alpha$ and weights of original data $\gamma_\textrm{O}$ during training. 
    Traditional data augmentation (yellow solid box in lower left) uses modest augmentation strength ($\alpha\shortequals0.05, 1$) with no original examples for training ($\gamma_{\textrm{O}}\shortequals0$).
    The MTV data augmentation approach (green dashed box) suggests substantive weighting of original examples (e.g., $\gamma_{\textrm{O}}\shortequals0.5$) which allows for much stronger augmentation (e.g., $\alpha\geq 0.3$). 
    }
    \label{fig:heatmap_avg}
\end{figure}

\subsection{Choosing $\gamma_{\textrm{O}}$ and $\gamma_{\textrm{aug}}$ weighting}
As our experiments so far have used the MTV with balanced weighting of original and augmented data ($\gamma_\textrm{O}\shortequals\gamma_\textrm{aug}\shortequals0.5$), in this section we explore different weightings of $\gamma_{\textrm{O}}$ and $\gamma_{\textrm{aug}}$. 
Figure \ref{fig:heatmap_avg} shows these results averaged over all three datasets and all four augmentation techniques.
Traditional data augmentation, which uses modest augmentation strength (e.g., $\alpha \in \{0.05, 0.1\}$) and does not train on original data ($\gamma_\textrm{O}\shortequals0.0$), achieves reasonable performance gains.
As expected, when stronger augmentations were applied (e.g., $\alpha\geq 0.4$), training with only augmented data hurts performance. 
When training on both augmented and original data, however, performance improved with stronger augmentation and remained robust for varying augmentation strengths $0.2\leq\alpha\leq0.5$ and original data weights $0.3\leq\gamma_{\textrm{O}}\leq0.7$.

\section{Further Related Work}
\vspace{-1.3mm}
Prior work on data augmentation, to our knowledge, generally follows the traditional data augmentation framework.
In addition to the methods mentioned in $\S\ref{sec:traditional}$, 
Xie et al.~\shortcite{DBLP:conf/iclr/XieWLLNJN17} replaced words with samples from the unigram frequency distribution;
Yu et al.~\shortcite{DBLP:conf/iclr/YuDLZ00L18} translated English sentences to French and back to English (backtranslation); 
and Kobayashi~\shortcite{kobayashi-2018-contextual} replaced words with other words based on a language model.
All these methods could potentially be formulated in the MTV.

Some prior work has also drawn connections between seeing data augmentation as multiple tasks. 
Similar to how we optimize augmented data as a separate task, \citet{DBLP:conf/icml/MeyersonM18} created fake tasks by using multiple distinct decoders to train a shared structure to solve the same problem in different ways.
In machine translation, \citet{sennrich-etal-2016-improving} used monolingual training examples as parallel examples with an empty source side, noting that their setup could be seen as multi-task learning with the tasks as translation with known sources and language modeling with unknown sources.
Compared with these papers that create multiple tasks in very specialized scenarios, the multi-task view that we have presented here can be used for any type of text data augmentation.

To be clear, our study is not the first to mix original and augmented data in training.
For instance, \citet{wang-yang-2015-thats} use a ratio of 1:5 original to augmented examples, but this weight of original data is much smaller than the $0.3\leq\gamma_{\textrm{O}}\leq0.7$ that we advocate for.  
\citet{sennrich-etal-2016-improving} also include original data when training with back-translation augmentation, but the given ratios of original and augmented data they use appear to dictated by the speed of their back-translation models rather than an intentionally-motivated design choice.
We see our work as the first to explicitly formulate the MTV, advocate for a joint optimization function, and comprehensively explore its implications on common text augmentation techniques.

As a limitation, our study has focused on label-preserving augmentation techniques, and our line of reasoning may not apply when augmentation techniques intentionally change the label.
Moreover, we have only studied text classification with simple models using task-agnostic augmentation techniques.
Future work in this direction could experiment with larger-scale models or study task-specific augmentation.

\section{Conclusions}
\vspace{-1.3mm}
We have proposed a multi-task view that gives both original and augmented examples substantial weight during training, contrasting prior work that performs stochastic data augmentation and ignores original training data.
For four common augmentation techniques, we found experimentally that this alternative view allows for stronger levels of augmentation, which in turn leads to better and more robust performance than traditional augmentation. 
We hope our paper inspires future work using text data augmentation to think more explicitly about how much augmented examples resemble original data and consider substantive weighting of original data when using data augmentation to improve model performance.

To close, we leave the enthusiastic reader with one last thought.
Most existing text data augmentation techniques have obediently followed the paradigm from computer vision of generating augmented examples that are similar to the original data.
Who's to say that's how data augmentation ought to work in NLP?
In this paper, we've shown how to search for relative freedom from this constraint, simply by taking a different view of the underlying assumptions. 
Now, a bigger question arises on the horizon---what new text augmentation techniques are unlocked when augmented data are not forced to resemble the original?

\bibliography{anthology,eacl2021}
\bibliographystyle{acl_natbib}

\end{document}